\documentclass[10pt,journal,compsoc]{IEEEtran}
\usepackage{amssymb,amsmath}
\usepackage{balance}
\usepackage{epsfig}
\usepackage{algorithm} 
\usepackage{algorithmic} 
\usepackage{booktabs}
\usepackage{graphicx,subfigure}
\usepackage{graphics}
\usepackage{array}
\usepackage{multirow}
\usepackage{flushend}
\usepackage{color}

\ifCLASSOPTIONcompsoc
  \usepackage[nocompress]{cite}
\else
  \usepackage{cite}
\fi

\ifCLASSINFOpdf
\else
\fi

\begin{document}
%
\title{Spatiotemporal Co-attention Recurrent Neural Networks for Human-Skeleton Motion Prediction}
%
%
%
%

	\author{Xiangbo Shu, Liyan Zhang, Guo-Jun Qi, Wei Liu, and Jinhui Tang
	\thanks{\em X. Shu, and J. Tang are with the School of Computer Science and Engineering, Nanjing
		University of Science and Technology, Nanjing 210094, China. E-mail:$\{$shuxb, jinhuitang$\}$@njust.edu.cn.}
	\thanks{\em L. Zhang is with the College of Computer Science and Technology, Nanjing University of Aeronautics and Astronautics, Nanjing 210016, China. E-mail: zhangliyan@nuaa.edu.cn.}
	\thanks{\em G.-J. Qi is with the Huawei Cloud, Bellevue, WA 98004, USA. E-mail: guojun.qi@huawei.com.}
	\thanks{\em W. Liu is with the Computer Vision Group, Tencent AI Lab, Shenzhen
		518000, China. E-mail: wliu@ee.columbia.edu.}
}

%
%

\markboth{SUBMISSION~FOR~Journal, 2019}%
{SUBMISSION~FOR~JournalE, 2019}
%



\IEEEtitleabstractindextext{%
\begin{abstract}
	Human motion prediction aims to generate future motions based on the observed human motions. Witnessing the success
	of Recurrent Neural Networks (RNN) in modeling the sequential data, recent works utilize RNN to model human-skeleton motion on the observed motion sequence and predict future human motions. However, these methods did not consider the existence of the spatial coherence among joints and the temporal evolution among skeletons, which reflects the crucial characteristics of human motion in spatiotemporal space. To this end, we propose a novel Skeleton-joint Co-attention Recurrent Neural Networks (SC-RNN) to capture the spatial coherence among joints, and the temporal evolution among skeletons simultaneously on a skeleton-joint co-attention feature map in spatiotemporal space. First, a skeleton-joint feature map is constructed as the representation of the observed motion sequence. Second, we design a new Skeleton-joint Co-Attention (SCA) mechanism to dynamically learn a skeleton-joint co-attention feature map of this skeleton-joint feature map, which can refine the useful observed motion information to predict one future motion. Third, a variant of GRU embedded with SCA collaboratively models the human-skeleton motion and human-joint motion in spatiotemporal space by regarding the skeleton-joint co-attention feature map as the motion context. Experimental results on human motion prediction demonstrate the proposed method outperforms the related methods.
\end{abstract}

}

\maketitle

\IEEEdisplaynontitleabstractindextext

\IEEEpeerreviewmaketitle

	\section{Introduction}
\IEEEPARstart{W}{ith} the development of the MOCAP devices\footnote{https://en.wikipedia.org/wiki/Motion\underline{~~}capture}, such as Kinect, PS Move and Vicon, capturing human-skeleton motions becomes easy and convenient. At present, more and more MOCAP-based products are emerging in various application areas, such as somatic game, Virtual Reality (VR), and human-machine interactions, etc. One of the key technologies behind these products are the human-skeleton motion analysis. Human-skeleton motion analysis is an interesting topic in the computer vision area, generally including motion synthesis, motion prediction, pose estimation and so on~\cite{martinez2017human,butepage2017deep,lehrmann2014efficient,presti20163d}. Besides, many researchers have also attempted to understand the human action, interaction and even activity in video by taking the human-skeleton motion as a crucial motion information~\cite{zhu2016co,du2015hierarchical,vemulapalli2014human,mahasseni2016regularizing}.			

\begin{figure}[!t]
	\vspace{4mm}
	\centering
	\includegraphics[scale=0.19]{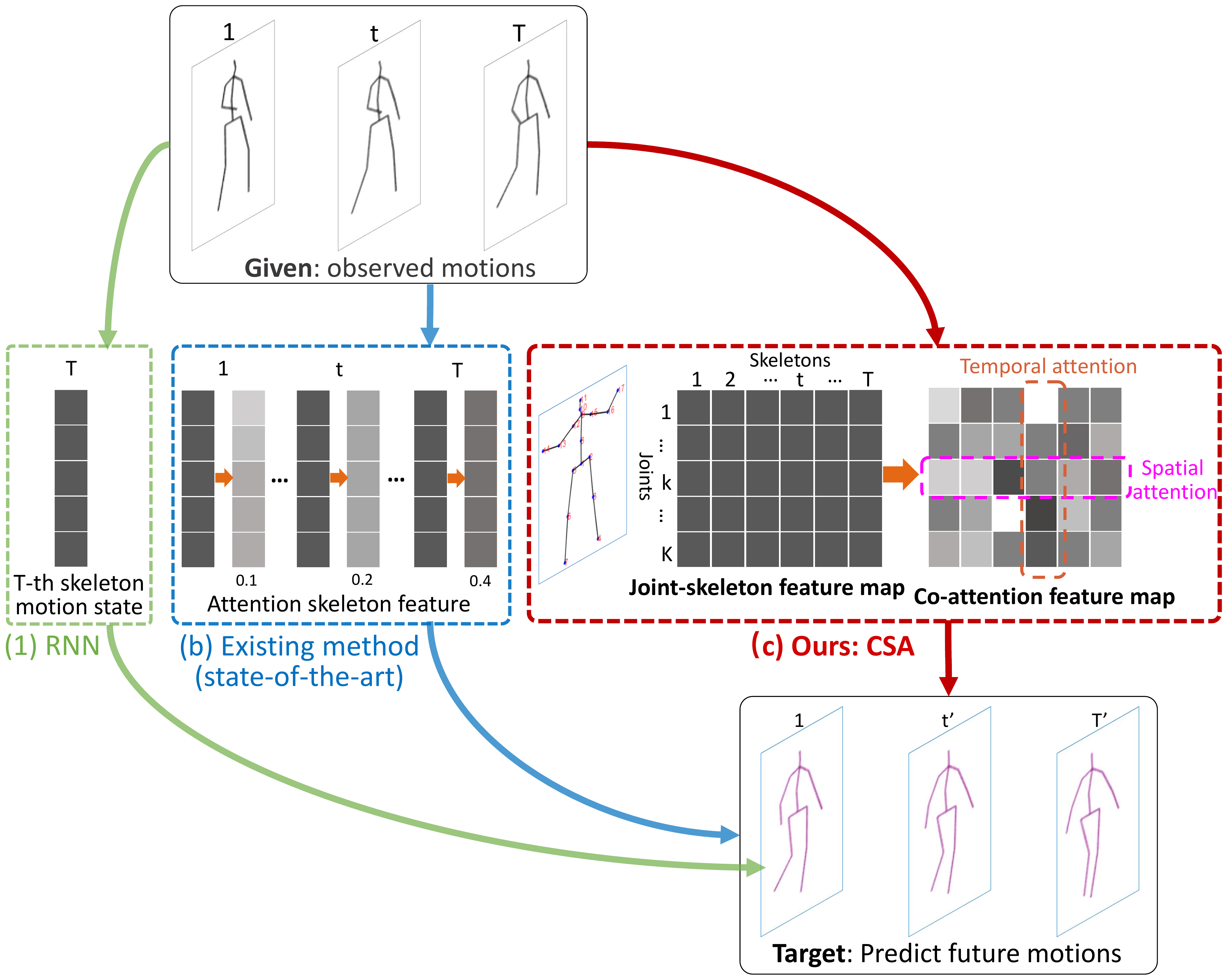}
	\caption{Idea of Skeleton-joint Co-Attnetion (SCA). Existing methods only learn the skeleton-attention feature vectors in temporal space, where all elements in a vector are assigned with the same attention factor. Our SCA dynamically learns a skeleton-joint co-attention feature map of human joints and skeletons in spatiotemporal space, where all elements in such feature map are assigned with different attention factors. SCA will be embedded into a variant of RNN, called Spatiotemporal Co-Attention RNN (SC-RNN) in this paper.}
	\label{idea}
\end{figure}

In this work, we focus on the human motion prediction task that aims to generate the future skeleton motions based on the observed skeleton motion sequence captured by the MOCAP. In the early stages, due to the sequentiality and nonlinearity of human motions, researchers utilized various probabilistic models, such as latent variable
model~\cite{urtasun2007modeling}, Markov model~\cite{lehrmann2014efficient,pavlovic2001learning}, generative model~\cite{taylor2007modeling}, to predict human motion based on a number of observed human motions. 
Recently, Recurrent Neural Networks (RNN)~\cite{yu2016video}, especially Gated Recurrent Unit (GRU)~\cite{chung2014empirical}, and Long Short-Term Memory (LSTM)~\cite{hochreiter1997long}, has shown remarkable performance in capturing the sequential information compared with traditional methods~\cite{shu2018hierarchical}, \cite{shu2017concurrence}, \cite{yan2018participation}. Therefore, recent RNN-based methods~\cite{fragkiadaki2015recurrent,jain2016structural,martinez2017human} used RNN to model the sequential motions over time for addressing the problem of human motion prediction.

In above RNN-based motion prediction methods, the direct propagation of motion information only happens between the consecutive frames. Although RNN can predict a satisfactory short-term motion, it weakly predicts one long-term motion only by the motion state updated in the previous frame~\cite{tang2018long}.  To address this problem, a straightforward way is that we can directly propagate all the observed motions into one future frame for predicting the long-term motion in RNN. In other words, we can set all the observed motions as a new motion context, which directly propagates into one future frame in RNN. However, not all observed motions equally contribute to the prediction of one specific future motion. Therefore, Tang et al.~\cite{tang2018long} presented a temporal attention mechanism (namely skeleton attention) that dynamically learns an attention factor for each observed motion based on their contribution to the prediction of one future motion. In this work, by regarding all the observed motions as the motion context, we utilize RNN with skeleton-attention mechanism to predict future motion. 

Moreover, most RNN-based motion prediction methods model human-skeleton motions only in temporal space, which ignore the spatial coherence among human joints in spatial space. For example, in the ``walking" action, the swing directions of left(right) arm and right(left) leg are usually consistent with each other in spatial space. Some previous action recognition methods have validated that exploring the spatial dependence among joints and the temporal dependence among skeletons can further improve the performance on human action recognition~\cite{wang2017modeling,liu2018skeleton}. Therefore, we consider exploring the spatial coherence among human joints and temporal evolution among human skeletons simultaneously via an attention mechanism in spatiotemporal space (not a single space). 

Motivated by this, we design a new {\bf Skeleton-joint Co-Attention (SCA)} mechanism to well capture the whole human motion information by learning the skeleton-joint co-attention factors of all observed motions in spatiotemporal space, as shown in Figure~\ref{idea}. We firstly construct a skeleton-joint feature map to represent all the observed motions, in which one row and column indicate a joint-type feature and skeleton-type feature of human motion in the spatial and temporal spaces, respectively. On the skeleton-joint feature map, SCA dynamically learns the skeleton-attention factors of human skeletons in temporal space, and the joint-attention factors of human joints in spatial space. At each time step, we can obtain a skeleton-joint co-attention feature map, which refines the useful observed motion corresponding to one future motion by the different attention factors. As we know, in conventional attention models, one feature vector is assigned with an attention factor in a single space, e.g., temporal space. Unlike conventional attention models, in SCA, each element in the skeleton-joint co-attention feature map is assigned with a skeleton-joint co-attention factor (obtained through multiplying a skeleton-attention factor by a joint attention factor) in multiple spaces (i.e., both spatial and temporal spaces) that measures its contribution to the prediction of one future motion in spatiotemporal space.

\begin{figure*}[!t]
	\vspace{3mm}
	\centering
	\includegraphics[scale=0.21000]{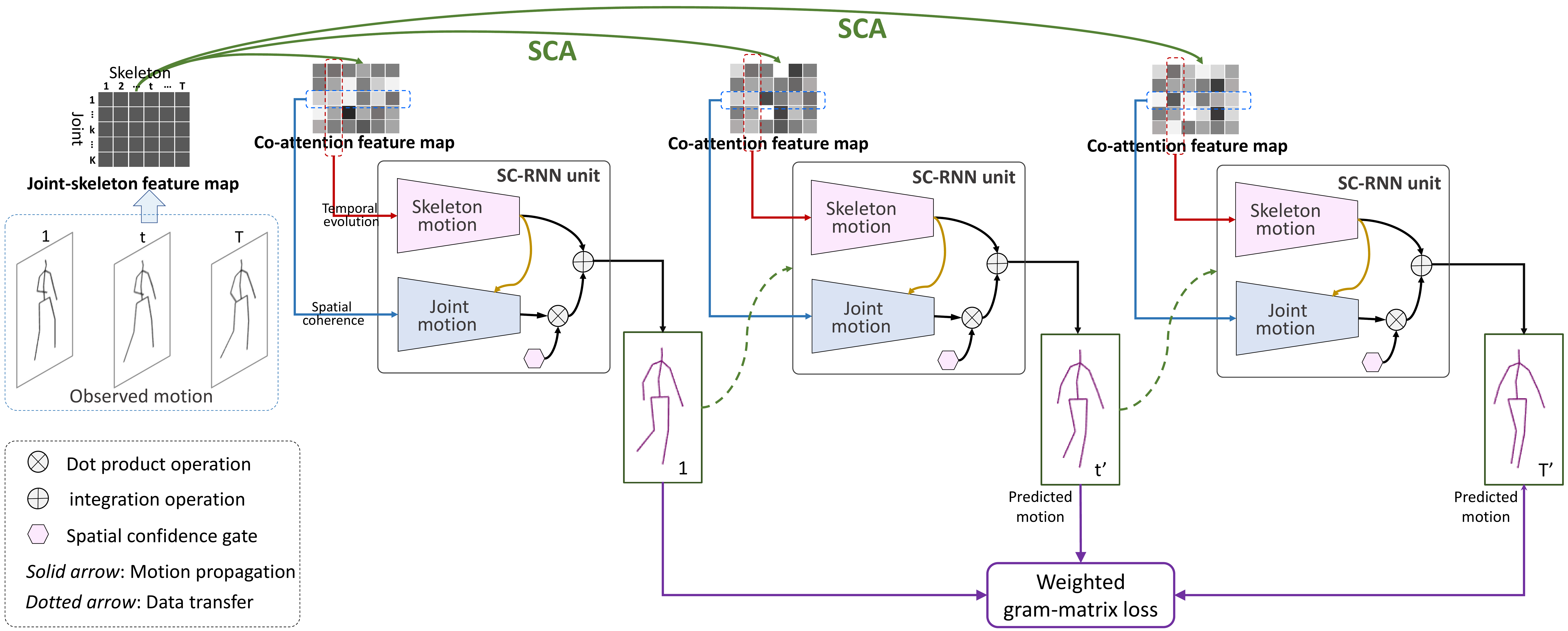}
	\vspace{0.5mm}
	\caption{Framework of the proposed SC-RNN in modeling human motions. At each time step, a co-attention feature map of all the observed motions is firstly learned by Skeleton-joint Co-Attention (SCA) in spatiotemporal space. CS-RNN models human-skeleton motions and human-joint motions in spatiotemporal space by regarding the skeleton-joint co-attention feature map as a new motion context. A new weighted gram-matrix loss is presented to learn SC-RNN in the training phase.}
	\label{framework}
\end{figure*}

Based on the designed SCA, we propose a novel {\bf Skeleton-joint Co-attention Recurrent Neural Networks (SC-RNN)} to simultaneously capture the spatial coherence among joints and the temporal evolution among skeletons  on a co-attention feature map in spatiotemporal space. Figure~\ref{framework} shows the training framework of SC-RNN. Specifically, SC-RNN models human-skeleton motions and human-joint motions simultaneously, by regarding the skeleton-joint co-attention feature map as a new motion context. And then, the predicted human-skeleton motions and human-joint motions are integrated to predict the future motion by a new spatial confidence gate at each time step. Finally, we propose a new weighted gram-matrix loss to measure the errors between the predicted motion and the ground-truth motion in the model learning.  In this work, we conduct experiments on human motion prediction to show the superiority of the proposed SC-RNN compared with the related methods. 

Over all, the main contributions of this work are summarized as follows:
	\begin{itemize}
	\item  We propose a novel Skeleton-joint Co-attention Recurrent Neural Network (SC-RNN) to simultaneously capture the spatial coherence among joints  and the temporal evolution among skeletons in spatiotemporal space.
	 \item We design a new Skeleton-joint Co-Attention (SCA) mechanism to dynamically refine the observed motion information over time by learning a skeleton-joint co-attention feature map on the observed human motion sequence. 
	 \item To learn SC-RNN, we propose a new weighted gram-matrix loss that exploits the correlations between two skeletons, which ensures the predicted skeletons with high correlation having the similar location. 
		\end{itemize}

	The rest of this paper is organized as follows. Section~\ref{RW} reviews some works related to skeleton-based action recognition, and human-skeleton motion prediction. Section~\ref{TPM} introduces the proposed method in details. Experiments are conducted in Section~\ref{EXP}, followed by the conclusions in Section~\ref{C}.

\vspace{3mm}
\section{Related Work}
\label{RW}
In this section, we briefly review some works related to the skeleton-based action recognition and human-skeleton motion prediction. 
\subsection{Skeleton-based Action Recognition}
Human action recognition is a long-standing research topic in computer vision and pattern recognition communities, which aims to enable a computer to automatically
understand the action performed by people~\cite{poppe2010survey,cheng2015advances,herath2016going}. In the task of action recognition, the input data are RGB image or skeleton of each frame. Based on different input data, action recognition can be mainly categorized two classes, i.e., RGB-based action recognition and skeleton-based action recognition. In RGB-based action recognition, the input features are the visual features of RGB-image in each frame~\cite{shu2017concurrence,veeriah2015differential,shu2018hierarchical,song2017end}, and sometimes they are attached the flow features~\cite{zhang2012spatio,simonyan2014two,ke2007spatio,feichtenhofer2016convolutional,varol2018long}. In skeleton-based action recognition, the input features are the positions of the skeleton ~\cite{liu2016spatio,wang2017modeling,wang2014learning,zhang2017view,rahmani2017learning,yan2018spatial}. We will mainly focus skeleton-based action recognition in this paper. 

As one of the earliest works, Yacoob and Black~\cite{yacoob1999parameterized} proposed to model the time-scale and time-shifted activity by Principal Component Analysis (PCA). Subsequently, some researchers proposed various models to learn representations of actions~\cite{rahmani2015learning,lillo2014discriminative,wang2014learning,shahroudy2016multimodal,weng2017spatio}. For example, since the tracked 3D positions of human joints may be wrong, Wang et al.~\cite{wang2014learning} proposed an ``actionlet" ensemble model to learn the representation of each action by capturing the intra-class variance. Shahroudy et al.~\cite{shahroudy2016multimodal} divided the human motion into several body parts, and proposed a joint sparse regression method to utilize the structured sparsity to model the features from a subset of the body parts. 

Recently, Deep neural Network has been employed to learn the more discriminative representations of actions\cite{hou2018skeleton,huang2017deep,yan2018spatial,shahroudy2018deep}. In particular, as an advantage for handling sequential data with variable length~\cite{sutskever2011generating,fragkiadaki2015recurrent}, Recurrent Neural Networks, including Gated Recurrent Unit, and Long Short-Term Memory (LSTM), has made impressive progress in skeleton-based action  recognition~\cite{du2015hierarchical,shi2017learning,liu2016spatio,zhang2017view,liu2017global,liu2018skeleton,wang2017modeling,zhu2016co}. One of the most representative works is that Du et al.~\cite{du2015hierarchical} divided a human body skeleton into  five  parts,  and  then fed  these  five  parts  into  five  RNNs.  As  the  number  of
layers increases, the outputs of multiple RNNs are hierarchically fused, and input to the following RNNs. Besides, Liu et al.~\cite{liu2016spatio,liu2018skeleton}  utilized LSTM  to model the temporal and spatial dependencies among  human motions for better understanding human action in the skeletal data. Likewise,  Wang et al.~\cite{wang2017modeling}  proposed  a  two-stream RNN (i.e., temporal RNN and spatial RNN) architecture to model temporal motions of skeletons over time and spatial relations among skeleton joints. These methods hold that all human joints contribute equally to the action. In fact, not all human joints are informative for action recognition and some irrelevant joints will bring in noise information. Therefore, Liu et al.~\cite{liu2017global} utilized an attention mechanism to capture the important human joints, and proposed a global context-aware attention LSTM to model human motion on the human joints with different attention factors.

\subsection{Human-Skeleton Motion Prediction}
Human-skeleton motion prediction aims to generate the future motions based on the observed human motions, which is becoming an attractive topic in the area of computer vision. It motivates a wide spectrum of applications, such as athletic performance analysis, surveillance, man–machine interfaces, somatic game,  Virtual Reality (VR), and so on~\cite{aggarwal1999human}. Generally, the human-skeleton motion data is captured by a MOCAP device, e.g., Kinect, Vicon. At present, there are some human-skeleton motion datasets released on websites, such as H3.6m mocap dataset (H3.6M)~\cite{ionescu2014human3}, CMU mocap database\footnote{http://mocap.cs.cmu.edu/}, 
In particular, the H3.6M mocap dataset is the largest-Scale and available human skeleton dataset for evaluating human skeleton prediction. 

In the early stage, since human motion is sequential and nonliner over time, some researchers utilized various probabilistic models, e.g., latent variable
model~\cite{urtasun2007modeling}, Markov model~\cite{lehrmann2014efficient,pavlovic2001learning}, generative model~\cite{taylor2007modeling}, and Gaussian process dynamical model~\cite{wang2008gaussian}, to predict the future human motion sequence based on the historical human motions. For example, Taylor et al.~\cite{taylor2007modeling} proposed non-linear generative model to utilize an undirected model with binary latent variables and real-valued ``visible" variables to represent joint angles. Lehrmann et al.~\cite{lehrmann2014efficient} proposed a non-parametric and nonlinear Markov model instead of simple Markov models for flexibly inferring future motions. 

Recently, witnessing the powerful ability of RNN in the action recognition task, some researchers proposed various variants of RNN to model the human-skeleton motion based on an observed motion sequence~\cite{fragkiadaki2015recurrent,martinez2017human,jain2016structural,pavllo2018quaternet}, which outperformed conventional methods on the motion prediction task. Fragkiadaki et al.~\cite{fragkiadaki2015recurrent} first employ RNN model to address the problem of human motion predictions. The authors proposed an Encoder-Recurrent-Decoder (ERD) model to encode and decode human motions by incorporating nonlinear encoder and decoder networks before and after recurrent layers. However, ERD is prone to accumulate errors, and quickly produces unrealistic human motion, though an added noise scheduling can reduce such errors to some extend. Subsequently, Martinez et al.~\cite{martinez2017human} 
embedded the residual module in the single-layer GRU architecture to model first-order motion derivatives. One limitation is that these methods pay no attention to the dependence of some joints in the spatial space. In fact, two neighboring joints in a frame usually have the coherent motion direction. Based on this, Jain et al.~\cite{jain2016structural} extend a spatio-temporal graph to a structural RNN, in which each node denotes a human joint, and the adjacency of graph between two nodes (joints) describe the dependence of two joints. This motivates us to model human motions in both the spatial and temporal spaces. 

The above RNN-based human motion prediction methods use the last motion and the current input motion to decide the future motion, this may be bring in noise motion information when the last motion is a sudden motion that is not consistent with the previous motions, or the last motion is not inaccurate enough. Therefore, Tang et al.~\cite{tang2018long} considered using all historical motions instead of the last motion to predict the next motion in the RNN model. Specifically, they presented an attention mechanism to assign all useful historical motions with different attention factors based on their contributions to prediction of the next motion. Inspired by this, we further consider embedding the attention mechanism into RNN to model human motions in both of the spatial and temporal spaces.

\vspace{3mm}
\section{Methodology}
\label{TPM}
\subsection{Background and Idea} \label{Primarily}
Let $ {\bf x}_{t}$ denote a representation of the human-skeleton motion in frame $t$ and  $\{ {\bf x}_{t}\}_{t=1}^{T}$ denote an observed human-skeleton sequence $\mathcal{S}$, as shown in Table~\ref{tab_notation}, the goal of human-skeleton motion prediction is to generate the future human-skeleton motion $\{{{\bf \tilde x}}_{t'}\}_{t'=1}^{T'}$ close to the ground-truth skeleton motion $\{{\hat {\bf x}}_{t}\}_{t=1}^{T}$ as much as possible. Some conventional works~\cite{fragkiadaki2015recurrent,martinez2017human} utilized Recurrent Neural Network (RNN) to predict human motion ${\tilde {\bf x}}_{t'+1}$ at the future time step $(t'+1)$ by the last motion state ${\bf h}_{t'-1}$ and the last predicted motion ${\tilde {\bf x}}_{t}$, as follows,
\begin{equation} \label{eq1}
{\tilde {\bf x}}_{t'+1}:={\text {RNN}}({\tilde {\bf x}}_{t'},{\bf h}_{t'-1}),
\end{equation}
The final motion state and human motion ${\bf x}_{T}$ of observed human motions are denoted by ${\bf h}_{0}$ and ${\tilde {\bf x}}_{0}$ when predicting the skeleton motion ${\tilde {\bf x}}_{1}$.

If there is a sudden motion in the last time step, the normal motion information will be disturbed or some noise information will be brought in. Therefore, we can use all historical motion information to predict the future motion ${\tilde {\bf x}}_{t'+1}$ by RNN~\cite{tang2018long}, i.e., 
\begin{equation} \label{eq2.0}
{\tilde {\bf x}}_{t'+1}:={\text {RNN}}({\tilde {\bf x}}_{t'},{\text {\bf Attention}}(\{{\bf x}_{t}\}_{t=1}^{T})).
\end{equation}
where ${\text {\bf Attention}}(\cdot)$ indicates an attention mechanism that computes the attention factors of all historical skeletons in the temporal space. Different historical motions are weighted by the different attention factors due to their contributions to the prediction of future motion at time step $(t'+1)$. In the other words, the more important skeleton motion will be assigned the larger attention factor.

However, the learning model in~Eq.~\eqref{eq2.0} ignores the spatial coherence among joints in the spatial space. For example, two neighboring joints in the spatial space usually have the similar motion direction. Therefore, we design a new {\bf Spatiotemporal Co-Attention (SCA)} mechanism to compute the spatial-attention factors and the temporal-attention factors corresponding to joints in the spatial space and skeletons in the temporal space, respectively. The learning model of Eq.~\eqref{eq2.1} can be embedded with SCA as follows,
\begin{equation} \label{eq2.1}
{{\bf \tilde x}}_{t'+1}:={\text {RNN}}({\bf \tilde {\bf x}}_{t'},{\text {\bf SCA}}(\mathcal {S})).
\end{equation}

Based on this, we propose a novel {\bf Spatiotemporal Co-attention Recurrent Neural Networks (SC-RNN)} to capture the spatial coherence among joints and the temporal evolution among skeletons simultaneously by utilizing all useful historical motion information as the motion context. SC-RNN mainly consists of a Spatial Attention GRU, a Temporal Attention GRU, a spatial confidence gate, and a weighted gram matrix-based loss, as shown in Figure~\ref{framework}. The details of SC-RNN will be introduced in the following sections.


\begin{table}[t!]
	\vspace{3mm}
	\renewcommand{\arraystretch}{1.2}
	\centering
	\caption{Notations and definitions.}
	\label{tab_notation}
	{\scriptsize
		\begin{tabular}{ll}
			\hline
			Notation & Definition\\
			\hline
			$t\in\{{1,2,\cdots,T\}}$& Index of the observed motion sequence.\\
			$t'\in\{{1,2,\cdots,T'\}}$& Index of the future motion sequence.\\
			${\bf x}_{t}$& Observed skeleton sequence at historical time step $t$.\\
			${\tilde {\bf x}}_{t'}$& Predicted motion at future time step $t'$.\\
			${\hat {\bf x}}_{t'}$& Ground-truth motion at future time step $t'$.\\
			$k\in\{{1,2,\cdots,K\}}$& Index of the observed joint sequence.\\
			${\bf F}$ & Skeleton-joint feature map covering $\{1, \cdots, T\}$.\\
			${\bf F}^{t'}_a$ & Skeleton-attention feature map covering $\{1, \cdots, T, t'\}$.\\
			${\bf F^{t'}}$ & Skeleton-joint feature map covering $\{1, \cdots, T, t'\}$.\\
			${{\bf F}_{co}^{t'}}$ & Skeleton-joint co-attention feature map covering $\{1, \cdots, T, t'\}$.\\
			\hline
		\end{tabular}
	}
\end{table}
\subsection{Skeleton-joint Co-attention Attention (SCA)} 
In this work, a human motion ${\bf x}_t$ at time step $t$ is represented by the human skeleton at this time step by concatenating the 3D location parameters of all $K$ joints in the angle axis. For one observed human motion sequence $\mathcal{S}=\{{\bf x}_t\in \mathbb{R}^{d}\}_{t=1}^{T}$, we can define a {\bf skeleton-joint feature map} ${\bf F}\in{\mathbb{R}}^{d\times T}$ (where $d=3\times K$) covering all the historical time steps $\{1, \cdots, T\}$, as follows,
	\begin{equation} \label{eq1}
	\begin{aligned}[l]
	{\bf F}_{(:,j)}\triangleq {\bf x}_{j}, ~ j=1,2,\cdots, T,
	\end{aligned}
	\end{equation}
	where ${\bf F}_{(:,j)}$ denotes the $j$-th column of ${\bf F}$. One row and column of ${\bf F}$ indicates a joint-type feature and skeleton-type feature of human motion in the spatial and temporal spaces, respectively.

As mentioned before, one long-term motion at the future time step can be well predicted by the current motion information, as well as all the observed motion information (i.e., motion context). However, not all observed skeleton motions equally contribute to the prediction of the motion at one future time step. If one observed motion is considerably consistent with one future motion, it could contribute a lot to predicting this future motion, and should be integrated into the motion context with a larger degree. Therefore, how to measure the consistency between an observed motion ${\bf x}_t$ and the future motion ${\bf \tilde{x}}_{t'}$ is the key. Inspired by the attention models in previous works~\cite{ramanathan2016detecting,tang2018long}, we adopt an attention mechanism to measure the consistency between the observed skeleton motion ${{\bf F}}_{(:,j)}$ and the predicted skeleton-motion state ${\bf h}_{t'-1}\in \mathbb{R}^d$ at time step $(t'-1)$, since ${\bf \tilde{x}}_{t'}$ is unknown.
Specifically, the skeleton-attention module of Skeleton-joint Co-Attention (SCA) in temporal space is formulated as follows,
\begin{equation}
	\beta^{e}_{j}={{\bf w}_e}^{\text T}\cdot  \varphi ({\bf U}_{eh}  {\bf h}_{t'-1}+{\bf U}_{ef} {\bf F}_{(;,j)}+{\bf b}_e);
\end{equation}
\begin{equation} \label{eq3}
	\begin{aligned}
		\{{{\alpha}}^e_{j}\}_{j=1}^T = {\text {softmax}}(\{\beta^{e}_{j}\}_{j=1}^T,\tau_1),\\
	\end{aligned}
\end{equation}
where $\tau_1$ is a temperature parameter of the softmax function~\cite{ramanathan2016detecting}, $\varphi$ is  a  hyperbolic  tangent, i.e.,  ${\text {tanh}}(\cdot)$, the superscript notation ${\text{T}}$ is the transposition operator of a vector/matrix, ${\bf U}_{e*}$ is the weight
matrix, ${\bf {w}}_e$ is a weight vector, and $b_e$ is a bias vector.

\begin{figure*}[!t]
	\centering
	\includegraphics[scale=0.29]{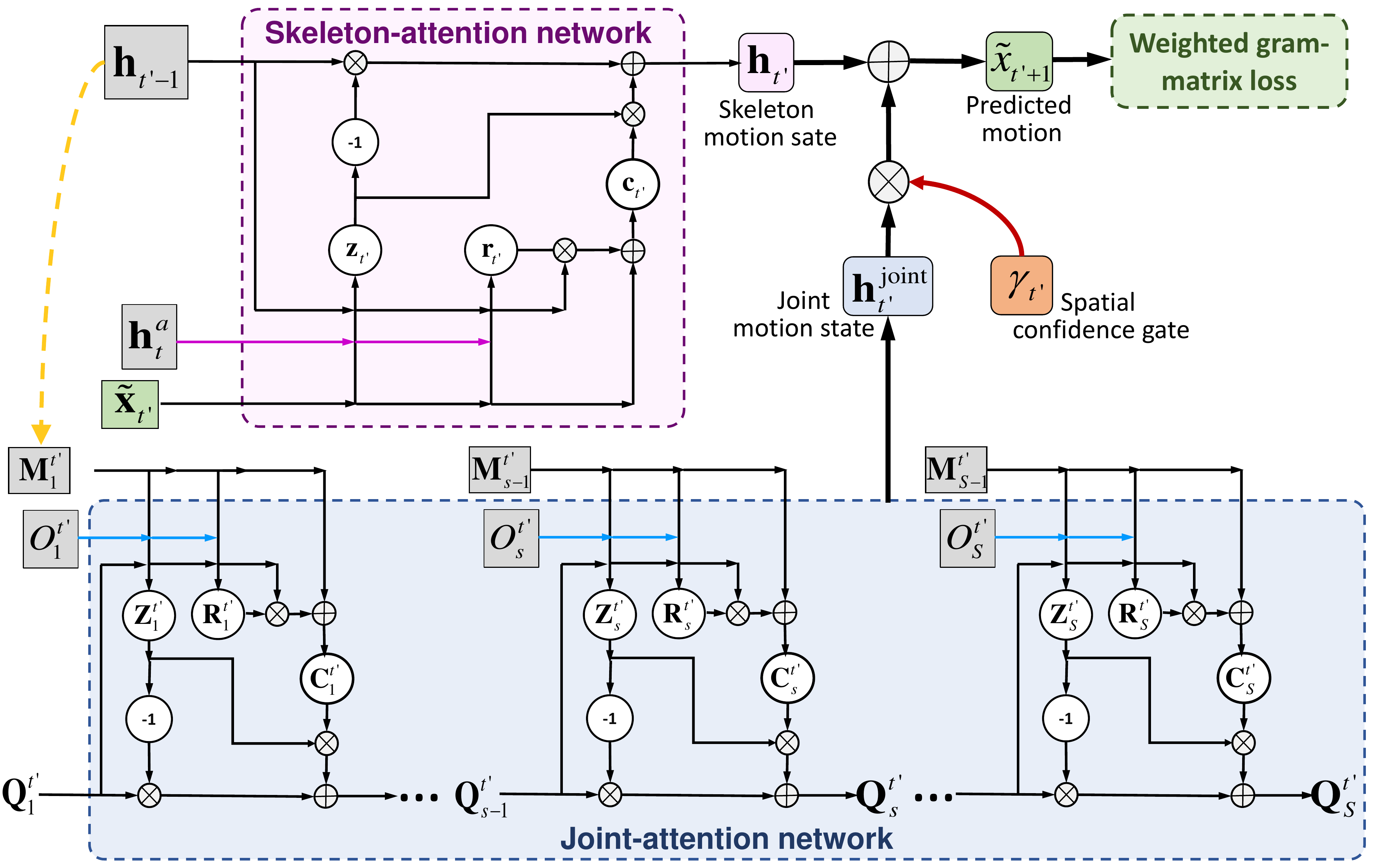}
	\caption{Architecture of SC-GRU (a GRU version of the proposed SC-RNN) at time step $t'$. SC-RNN mainly consists of the skeleton-attention network and joint-attention network, which simultaneously model the skeleton motion and joint motion.}
	\label{architecture}
\end{figure*}

After obtaining the skeleton-attention factor $\alpha_j^e$, the {\bf skeleton-attention feature map} ${\bf F}_a^{t'}$ and skeleton-attention feature ${\bf h}_a^{t'}$ (i.e., the {\bf skeleton motion context}) at time step $t'$ can be calculated as follows,
\begin{equation} \label{eq3}
	\begin{aligned}
		{{\bf F}_a^{t'}}_{(:,j)}=\alpha_j^e \cdot {\bf F}_{(:,j)};\\
		{\bf h}_a^{t'}= \frac{1}{T}{\sum}_{j=1}^{T}{{\bf F}_a^{t'}}_{(:,j)}.\\
	\end{aligned}
\end{equation}
By integrating the skeleton motion context ${\bf h}_a$, a new {\bf skeleton-attention network} is formulated as follows,
\begin{equation} \label{eq4}
	\begin{aligned}
		{\bf{z}}_{t'} = \sigma ({\bf{W}}_{zx} {\bf{\tilde x}}_{t'} +{\bf{W}}_{zh}  {\bf{h}}_{t'-1} +{\bf{W}}_{za} {{\bf{h}}_a^{t'}} + {\bf{b}}_z);
	\end{aligned}
\end{equation}
\begin{equation} \label{eq5}
	\begin{aligned}
		{\bf{r}}_{t'} = \sigma ({\bf{W}}_{rx} {\bf \tilde {x}}_{t'} +{\bf{W}}_{rh} {\bf{h}}_{t'-1}+{\bf{W}}_{ra} {{\bf{h}}_a^{t'}} + {\bf{b}}_r);
	\end{aligned}
\end{equation}
\begin{equation} \label{eq4}
	\begin{aligned}
		{\bf{c}}_{t'} = \varphi ({\bf{W}}_{cx} {\bf{ \tilde x}}_{t'} + {\bf{W}}_{ch} ({\bf{r}}_{t'}  \odot {\bf{h}}_{t'-1}) + {\bf{b}}_c);
	\end{aligned}
\end{equation}
\begin{equation} \label{eq14}
	\begin{aligned}
		\!\!{{\bf{h}}_{t'}} = (1\!-\!{\bf{z}}_{t'}) \odot{\bf h}_{t'-1}+  {\bf{z}}_{t'} \odot{\bf{c}}_{t'},~t'=1,\cdots,T',
	\end{aligned}
\end{equation}
where ${\bf z}_{t'}$, ${\bf r}_{t'}$, ${\bf c}_{t'}$, and ${\bf h}_{t'}\in \mathbb{R}^d$ are the update gate, reset gate, memory, and skeleton motion state at time step $t'$ respectively, $\sigma(\cdot)$ is a sigmoid function, $\odot$ denotes the element-wise product, $\varphi(\cdot)$ is a hyperbolic tangent $tanh(\cdot)$, ${\bf{W}}^e_{*x}$ and ${\bf{W}}^e_{*h}$ are weight matrices, and ${\bf b}_*$ is bias vector. Figure~\ref{architecture} shows the detailed architecture of skeleton-attention network.

Empirically, for the  $k$-th joint covering the time steps $\{1,\cdots,T,t'\}$, its motion is coherent to some other joints with different degrees in spatial space, Such as, the coherent motions of leg joint and hand joint in a ``walking". Therefore, it is crucial for measuring the coherence between the $k$-th joint motion and its neighboring joints. Similar to skeleton-attention module, we also adopt an attention mechanism to measure the coherence degree of the $k$-th joint motion and the $l$-th joint motion covering the time steps $\{1, \cdots, T, t'\}$, where $l=1,2,\cdots,K$, and $l\neq k$. Specifically, we first define a skeleton-joint feature feature map ${\bf F}^{t'}\in{\mathbb{R}}^{d\times(T+1)}$ covering the time steps $\{1, \cdots, T, t'\}$, i.e.,
\begin{equation} \label{eq2}
	\begin{aligned}
		{\bf F}^{t'}\triangleq[{\bf F},{\bf h}_{t'}],~ t'=1,2,\cdots, T',\\
	\end{aligned}
\end{equation}
where ${\bf h}_{t'}$ is obtained by the skeleton-attention network.
Subsequently, the joint-attention module of SCA can be expressed as below,
\begin{equation}
	\begin{aligned}
		\!\!{\bf \beta}^{k}_{l}={{\bf w}_c}^{\text T} \cdot \varphi ({\bf U}_{cb}\cdot{\bf M}^{t'}_{k}\cdot{\bf 1}+{\bf U}_{cm}\cdot{{\bf M}^{t'}_l}\cdot{\bf 1}+{\bf b}_l),\\
	\end{aligned}
\end{equation}
where ${\bf 1}\in \mathbb{R}^{T+1}$ denotes a vector with all elements being set to $1$, and we set $k_*=3\times(k-1)+*$, $l_*=3\times(l-1)+*$, $*=1,2,3$, ${\bf M}^{t'}_k=[{\bf F}^{t'}_{(k_1,:)};{\bf F}^{t'}_{(k_2,:)};{\bf F}^{t'}_{(k_3,:)}]$, and ${\bf M}^{t'}_l=[{\bf F}^{t'}_{(l_1,:)};{\bf F}^{t'}_{(l_2,:)};{\bf F}^{t'}_{(l_3,:)}]$ for brevity. A joint-attention factor $\alpha_l^{k}$ of the $l$-th joint corresponding to the $k$-th joint can be computed as below,
\begin{equation} \label{eq6}
	\begin{aligned}
		\{\alpha^{k}_{l}\}_{l=1,l\neq k}^{K} \!= {\text {softmax}}(\{\beta^{k}_{l}\}_{l=1,l\neq k}^{K},\tau_2),\\
	\end{aligned}
\end{equation}
where $\tau_2$ is a temperature parameter of softmax.

For the $k$-th joint, if we obtain $\{\alpha^{k}_{l}\}_{l=1,l\neq k}^{K}$, we can compute a {\bf skeleton-joint co-attention feature map} ${\bf F}_{co}^{t'}$ of ${\bf F}^{t'}$ of the skeleton attention feature map ${{\bf F}_a^{t'}}$ in spatiotemporal space, i.e.,
\begin{equation} \label{eq3}
	\begin{aligned}
		{{\bf F}_{co}^{t'}}_{(l_*,:)}=& \alpha^{k}_l \cdot [{{\bf F}_a^{t'}}_{{(l_*,1:T)}},{\bf F}^{t'}_{(l_*,{\text {end}})}], *=1,2,3,
	\end{aligned}
\end{equation}
where $l=1,\cdots,K$, and $l\neq k$. In this way, each element ${{\bf F}_{co}^{t'}}_{(i,j)}$ ($j=1,2,\cdots,T$) is assigned with a {\bf skeleton-joint attention factor}, where one skeleton-joint attention factor is obtained through multiplying a skeleton-attention factor by a joint attention factor. Then the {\bf co-attention motion context} ${\bf O}^{t'}_{k}$ of the $k$-th joint covering the time steps $\{1,2,\cdots,T,t'\}$ can be calculated as follows
\begin{equation} \label{eq3}
	\begin{aligned}
		{\bf q}^{*'}_{k}\triangleq& \frac{1}{K-1}\sum_{l=1,l\neq k}^{K}{{\bf F}_{co}^{t'}}_{(3\times(l-1)+*,:)}, *=1,2,3,\\
		{\bf O}^{t'}_{k}\triangleq&[{\bf q}^{1'}_{k};{\bf q}^{2'}_{k};{\bf q}^{3'}_{k}].
	\end{aligned}
\end{equation}

\begin{figure}[!t]
	\small{
		\centering
		\subfigure[]
		{
			\includegraphics[scale=0.48]{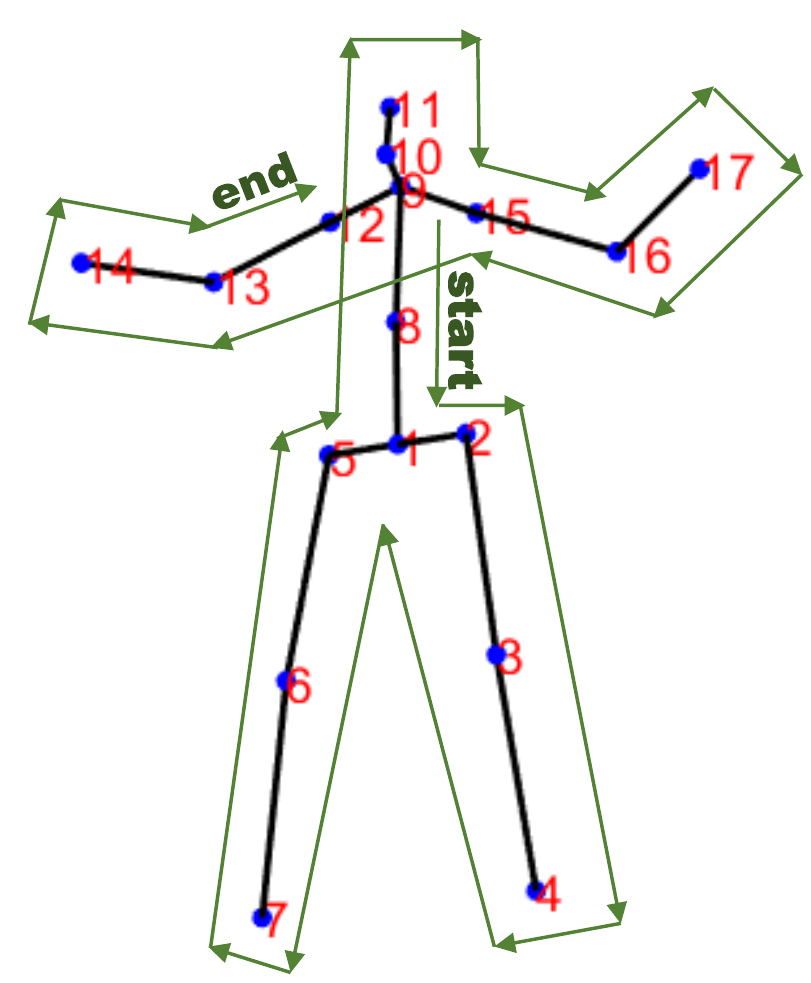}
			\label{fig_sequence_a}
		}~~~~
		\subfigure[]
		{
			\includegraphics[scale=0.48]{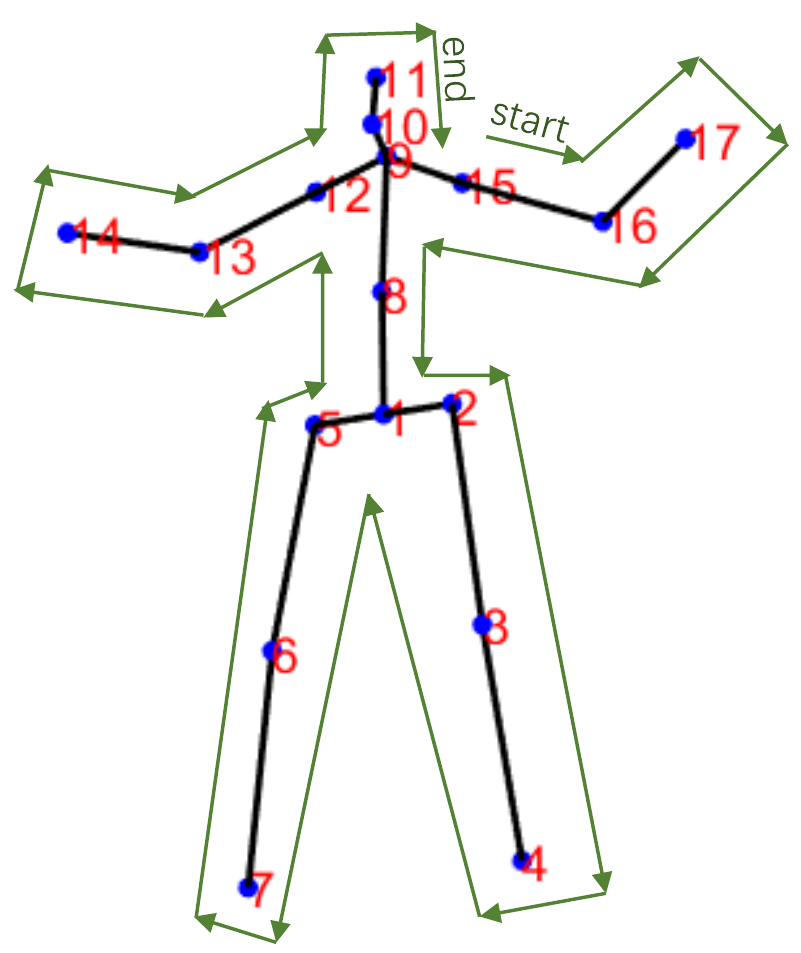}
			\label{fig_sequence_b}
		}
		\caption{Different orders for spatial arrangement. (a) and (b) represent
			the traveling-based and surrounding-based sequences, respectively.}
		\label{fig_sequence}
	}
\end{figure}

Similar to the skeleton-attention network, we can formulate a {\bf spatial-attention network} to learn the human-joint motion state ${\bf Q}^{t'}_s$ by integrating the co-attention motion context ${\bf O}^{t'}_{s}$  in a sequence $\{1,2,\cdots,s,\cdots,S\}$, 
\begin{equation} \label{eq17}
	\begin{aligned}
		\!\!{\bf{Z}}^{t'}_{s} = \sigma ({\bf{W}}_{zm} {\bf M}^{t'}_{s} +{\bf{W}}_{zq} {\bf{Q}}^{t'}_{s-1}+ {\bf{W}}_{zo}  {\bf O}^{t'}_{s} + {\bf{B}}_z);
	\end{aligned}
\end{equation}
\begin{equation} \label{eq18}
	\begin{aligned}
		\!\!{\bf{R}}^{t'}_{s} = \sigma ({\bf{W}}_{rm}  {\bf M}^{t'}_{s} +{\bf{W}}_{rq} {\bf{Q}}^{t'}_{s-1}+ {\bf{W}}_{ro} {\bf O}^{t'}_{s} + {\bf{B}}_r);
	\end{aligned}
\end{equation}
\begin{equation} \label{eq19}
	\begin{aligned}
		\!\!{\bf{C}}^{t'}_{s} = \varphi ({\bf{W}}_{cm} {\bf M}^{t'}_{s} + {\bf{W}}_{cq}  ({\bf{R}}^{t'}_{s}  \odot {\bf{Q}}^{t'}_{s-1}) + {\bf{B}}_c);
	\end{aligned}
\end{equation}
\begin{equation} \label{eq20}
	\begin{aligned}
		\!\!\!{{\bf{Q}}^{t'}_s} = (1-{\bf{Z}}^{t'}_{s}) \odot{\bf{Q}}^{t'}_{s-1}+  {{\bf{Z}}^{t'}_{s} \odot{\bf{C}}_s^{t'}},~s=1,\cdots\!,S,
	\end{aligned}
\end{equation}
where ${\bf M}^{t'}_s$, ${\bf Z}^{t'}_{s}$, ${\bf R}^{t'}_{s}$, ${\bf C}^{t'}_{s}$, and ${\bf Q}^{t'}_{s}$ are the input joint motion, update gate, reset gate, memory, and the joint motion state, respectively. $s$ is the index of in a pre-defined travel-based joint sequence $\Omega=\{1,2,\cdots, s,\cdots, S\}$~\cite{liu2018skeleton}. Intuitively, we can predefine the spatial joint sequence according the ID number of the joint and update all joints by Eq.~\eqref{eq17}$\sim$\eqref{eq20}, namely ID-based sequence. However, such joint sequence did not consider the joint structure. In some previous works~\cite{liu2018skeleton,cao2018skeleton}, the traveling-based and surrounding-based joint sequences are better to describe the human body structure, and achieve the better performance  compared with the traditional ID-based sequence in terms of action recognition. Therefore, we also adopt the traveling-based and surrounding-based sequences in this work. Specifically, the traveling-based and surrounding-based sequences are drawn in Figure~\ref{fig_sequence}, where the traveling-based and surrounding-base sequences are \{9, 8, 1, 2, 3, 4, 3, 2, 1, 5, 6, 7, 6, 5, 1, 8, 9, 10, 11, 10, 9, 15, 15, 17, 16, 15, 9, 12, 13, 14, 13, 12, 9\}, and \{9, 15, 16, 17, 16, 15, 9, 8, 1, 2, 3, 4, 3, 2, 1, 5, 6, 7, 6, 5, 1, 8, 9, 12, 13, 14, 13, 12, 9, 10, 11, 10, 9\}, respectively.

\subsection{Skeleton-joint Co-Attention GRU (SC-GRU)}
Till now, we can integrate the skeleton-attention network and joint-attention network into a Skeleton-joint Co-attention GRU (SC-GRU), namely a GRU version of the proposed SC-RNN\footnote{The other versions of SC-RNN (such as SC-RNN, SC-LSTM) can be easily formulated in the same way.}. Figure~\ref{architecture} shows the architecture of SC-GRU. In SC-GRU, at time step $t'$, we first obtain the skeleton motion state ${\bf h}_{t'}$ based on the human skeletons, and then update each joint to obtain ${\bf Q}^{t'}_k$ in spatiotemporal space. Finally, the predicted motion ${\bf \tilde {x}}_{t'+1}$ at time step $(t'+1)$ can be expressed as
\begin{equation} \label{eq21}
	{\bf \tilde {x}}_{t'+1}=\frac{1}{2}({\bf h}_{t'}+\gamma_{t'} \odot {\bf h}^{\textit {joint}}_{t'}),\\
\end{equation}
where ${\bf h}^{{\textit joint}}_{t'}=[{{{\bf Q}^{t'}_1}_{(:,{\text {end}})}};{{{\bf Q}_2^{t'}}_{(:,{\text {end}})}};\cdots;{{{\bf Q}_K^{t'}}_{(:,{\text {end}})}}]$. ${\bf \gamma}_{t'}$ is a spatial confidence gate at time step $t'$, which is activated by the difference between the predicted joint motion state ${\bf h}^{{\textit joint}}_{t'}$ and the input joint motion ${\bf m}^{{\textit joint}}_{t'}=[{{{\bf M}^{t'}_1}_{(:,{\text {end}})}};{{{\bf M}^{t'}_2}_{(:,{\text {end}})}};\cdots;{{{\bf M}^{t'}_K}_{(:,{\text {end}})}}]$, as follows
\begin{equation} 
	\label{eq22}
	\begin{aligned}
		{\bf f}^{t'}_h&=\varphi({\bf W}_{fh}{\bf h}^{{\textit joint}}_{t'}+{\bf b}_{h});\\
		{{\bf f}^{t'}_m}&=\varphi({\bf W}_{fm} {\bf m}_{t'}^{\textit joint}+{\bf b}_m);\\	
		{\gamma}_{t'}&=\chi({\bf f}^{t'}_h - {\bf f}^{t'}_m),
	\end{aligned}
\end{equation}
where $\chi({\theta})=\frac{1}{\exp(\rho {\theta}^{2})}$ is an element-wise activation function, $\rho$ is a parameter to control the bandwidth of the activation function. In Eq.~\eqref{eq22}, if the obtained joint motion state differs from the input joint motion state, the former has a lower confidence, and will be suppressed in the updating of the predicted motion ${\bf \tilde {x}}_{t'+1}$ in Eq.~\eqref{eq21}. 

\subsection{Weighted gram-matrix loss}
On the top of the SC-GRU, there is a loss function to measure the difference between the predicted motion and the ground-truth motion. Conventional methods~\cite{fragkiadaki2015recurrent,jain2016structural,martinez2017human} usually adopt Mean Square Error (MSE) as the loss function to minimize the errors between the predicted motion and ground truth in the model learning. However, MSE treats all predicted motions independently, which ignores the relation between motions at different time steps, and causes motion inconsistency to some extent. Recently, a gram-matrix loss instead of MSE is proposed to produce highly correlated human motion in temporal space~\cite{tang2018long}. However, it only considers the correlation between motions in two consecutive frames. Therefore, we extend the gram-matrix loss to a new weighted gram-matrix loss, as follows  
\begin{equation} 
	\label{eq23}
	\begin{aligned}
		\mathcal{L}=\frac{1}{T}{\sum}_{t'=1}^{T'}\left\| \left(\mathcal{G}\left({\bf{\tilde x}}_{t'}\right)- \mathcal{G}\left({\bf{\hat x}}_{t'}\right)\right) \right\|_F^2,
	\end{aligned}
\end{equation}
where $\mathcal{G}\left({\bf{\tilde x}}_{t'}\right)$ and $\mathcal{G}\left({\bf{\hat x}}_{t'}\right)$ are gram matrices corresponding to ${\bf{\tilde x}}_{t'}$ and ${\bf{\hat x}}_{t'}$, respectively, defined as
\begin{equation} 
	\label{eq21.0}
	\begin{aligned}
		\!\!\mathcal{G}\!\left({\bf{\tilde x}}_{t'}\!\right)\!\triangleq&[{\bf I}_{1,t'}{\bf{\tilde x}}_{1};{\bf I}_{2,t'}{\bf{\tilde x}}_{2};\cdots;{\bf I}_{t'-1,t'}{\bf{\tilde x}}_{t'-1};{\bf{\tilde x}}_{t'}]\\
		&\times [{\bf I}_{1,t'}{\bf{\tilde x}}_{1};{\bf I}_{2,t'}{\bf{\tilde x}}_{2};\cdots;{\bf I}_{t'-1,t'}{\bf{\tilde x}}_{t'-1};{\bf{\tilde x}}_{t'}]^{\text T};
	\end{aligned}
\end{equation}
\begin{equation} 
\label{eq21.1}
\begin{aligned}
\!\!\mathcal{G}\!\left({\bf{\hat x}}_{t'}\!\right)\!\triangleq&[{\bf I}_{1,t'}{\bf{\hat x}}_{1};{\bf I}_{2,t'}{\bf{\hat x}}_{2};\cdots;{\bf I}_{t'-1,t'}{\bf{\hat x}}_{t'-1};{\bf{\hat x}}_{t'}]\\
&\times [{\bf I}_{1,t'}{\bf{\hat x}}_{1};{\bf I}_{2,t'}{\bf{\hat x}}_{2};\cdots;{\bf I}_{t'-1,t'}{\bf{\hat x}}_{t'-1};{\bf{\hat x}}_{t'}]^{\text T}.
\end{aligned}
\end{equation}
In Eq.~\eqref{eq21.0} and~\eqref{eq21.1}, ${\bf I}\in{\mathbb R} ^{T' \times T'}$ is a coefficient matrix to measure the correction between two skeleton motions at different time steps, i.e.,
\begin{equation}
	\label{eq25}
	{\bf{I}}_{i,j}=exp
	(
	-\dfrac{||{\bf{\hat x}}_{i}-{\bf{\hat x}}_{j}||_2^2}{\tau^2}
	),~i,j=1,2,\cdots,T',
\end{equation}
where $\tau$ is a parameter of the RBF kernel. 

\vspace{3mm}
\section{Experiments}
We conduct experiments on human motion prediction to evaluate the performance of the proposed SC-RNN compared with the state-of-the-art methods.
\label{EXP}
\subsection{Dataset}
Following previous works~\cite{tang2018long,martinez2017human}, we select the H3.6m mocap dataset (H3.6M)~\cite{ionescu2014human3} to evaluate the performance of the proposed SC-RNN in the experiments. 
H3.6M is the largest available human-skeleton motion dataset captured, which is appropriate for training deep network with a huge number of parameters. It includes 15 actions are performed by seven different actors, i.e., ``directions", ``discussion", ``eating", ``greeting", ``phoning", ``posing", ``purchases", ``sitting", ``sittingdown", ``smoking", ``takingphoto", ``waiting", ``walking", ``walkingdog", and ``walkingtogether". 
	Some actions are periodic actions, e.g., ``walking", and the other ones are non-periodic, e.g., ``taking photo". For each action,  actor, and camera viewpoint, there are two video sequences, each of which has $3000\sim 5000$ frames. Each human-skeleton sequence records the 3D locations of 32 joints at each frame. Since many of joints are very close in the 3D angle axis, we select $17$ representative joints of each skeleton, which provide enough details of human motion.

\subsection{Experimental Setting}
For fair comparison, following setting in~\cite{fragkiadaki2015recurrent,jain2016structural,martinez2017human}, five actors' motions are selected for testing, while the rest ones for training. Each human joint is represented by a three-dimensional location vector in the 3D angle axis~\cite{taylor2007modeling}. Thus a human skeleton at one time step is represented by concatenating all locations of all joints. For each human-skeleton sequence, we take two seconds of motion as the observed ones to predict the future $400$ms and $1000$ms motion for short-term and long-term motion prediction, respectively. 

Following the evaluation in previous works ~\cite{jain2016structural,martinez2017human,tang2018long}, we adopt the Mean Angle Error (MAE) as the evaluation criteria, which measure the Euclidean distance between
the predicted joints and its ground-truth in the 3D angle axis.

In the configuration of SC-RNN, the number of memory cell nodes is set to $1024$. We use PyTorch toolbox and a NVIDIA Tesla K20 GPU to run the experiments. The learning rate, decay rate, momentum and batch size are set to $0.5\times10^{-3}$, $0.95$, $0.9$ and $32$, respectively. 

\begin{table} [t]
	\centering
	\caption{Mean Angle Error obtained by SC-RNN with different sequence strategies.}
	\label{table_sequence}
	\begin{center}
		\begin{tabular}{c|c|c|c|c}
			\hline\hline
			\hspace{-0.85em}\multirow{2}{*}{Strategies} &\multicolumn{2}{c}{Short-term } &\multicolumn{2}{c}{Long-term}\\
			\cline{2-5} &80ms   & 400ms  &   720ms  &     1000ms\\
			\hline
			ID-based sequence &0.40 & 1.06& 1.20 & 1.50\\
			Surrounding-based sequence &0.39 &{\bf 1.03} &1.18& 1.48\\
			Traveling-based sequence &{\bf 0.37} &1.04 &{\bf 1.17} & {\bf 1.46}\\
			\hline\hline
		\end{tabular}
	\end{center}
\end{table}

\begin{table*} [t]
	\vspace{3mm}
	\centering
				\caption{Performance comparison among different methods in terms of short-term and long-term human motion prediction for all 15 actions on the H3.6m
					dataset.}
{
		\begin{center}
			\begin{tabular}{l|c|c|c|c|c|c|c|c}
				\hline\hline
				\multirow{2}{*}{Methods} &\multicolumn{4}{c|}{Short-term motion prediction} &\multicolumn{4}{c}{Long-term motion prediction}\\
				\cline{2-9}
				& 80ms  & 160ms  &320ms &400ms   &   560ms&     640ms&     720ms &    1000ms\\
				\hline
				
				ERD~\cite{fragkiadaki2015recurrent}   & 0.60  & 0.85  & 1.15  & 1.36  & 1.54  & 1.63  & 1.70  & 1.80  \\
				LSTM-3LR~\cite{fragkiadaki2015recurrent} & 0.56  & 0.80  & 1.11  & 1.33  & 1.49  & 1.58  & 1.66  & 1.75  \\
				Res-GRU~\cite{martinez2017human}    & 0.40  & 0.70  & 1.06  & 1.18  & 1.37  & 1.43  & 1.53  & 1.63  \\
				MHU~\cite{tang2018long}   & 0.37  & 0.65  & 0.97  & 1.10  & 1.26  & 1.35  & 1.43  & 1.54  \\	
				\hline			
				SC-RNN w/o SCA & {0.49} & 0.72 &0.97 &1.20 &1.29& 1.40& 1.46&1.52 \\
				SC-RNN w/o skeleton attention & 0.36 & {\bf 0.60} &{{0.92}} &1.06 &1.24 & {{1.35}} &1.41 & 1.51\\
				SC-RNN w/o joint attention & {\bf 0.36}  &0.59 & 0.95 & {{1.03}} & {{1.20}} &1.38 &{\bf {1.34}} &{{1.49}}\\\hline
				SC-RNN &  0.37  & {{0.62}}  & {\bf 0.92}  & {\bf 1.04}  & {\bf 1.17}  & {\bf 1.25}  & {{1.36}}  & {\bf 1.46}\\
				\hline\hline
			\end{tabular}
			\label{table_results}
		\end{center}
	}
\end{table*}

In SA-GRU, we introduced traveling-based sequence and surrounding-based sequences, as shown in Figure~\ref{fig_sequence}. These two sequences have shown the competitive performance in the task of human action recognition~\cite{liu2018skeleton,cao2018skeleton}. For the task of human motion prediction, we firstly  explore the better sequence strategy by conducting the experiments based on three sequences, namely ID-based sequence, traveling-based sequence and surrounding-based sequences. The compared results are reported in Table~\ref{table_sequence}. It can be seen that the SC-RNN with traveling-based sequence achieves the best performance. Since two neighboring joints have the consistent motion direction, the adjacency information between any the two neighboring joints in the traveling-based will be better maintained. In the following experiments, SC-RNN is selected the traveling-based sequence by default.

\subsection{Ablation studies}
To illustrate the novelty of SC-RNN, we conduct ablation studies between SC-RNN and baseline methods. 1) {\bf SC-RNN w/o SCA}: it throws away the Skeleton-joint Co-attention Attention (SCA) in SC-RNN. 2) {\bf SC-RNN w/o skeleton attention}: it throws away the skeleton-attention module, namely $\{\alpha_j^e\}_{j=1}^T$ are setting to $1$. 3) {\bf SC-RNN w/o joint attention}: it throws away the joint-attention module, namely $\{\alpha_l^k\}_{l=1,l\neq k}^{K}$ are setting to $1$. 

Table~\ref{table_results} shows the average MAE of the proposed SC-RNN compared with the baselines on all 15 actions in the H3.6M dataset. we can see that SC-RNN outperforms all baseline methods. Specifically, ``SC-RNN w/o skeleton attention" and ``SC-RNN w/o joint attention" improve performance of ``SC-RNN w/o SCA" without attention mechanism to some extent. In particular, the MAE achieved by ``SC-RNN w/o skeleton attention" is comparable to ``SC-RNN w/o joint attention", where both of them adopt the attention mechanism in a single space. SC-RNN with SCA in spatiotemporal space outperforms ``SC-RNN w/o skeleton attention" and ``SC-RNN w/o joint attention" embedded with the ``single-attention" module in a single space. Overall, it can be concluded that SCA mechanism is more effective than a ``single-attention" mechanism in modeling human motions.

	\begin{table*} [t]
		\vspace{3mm}
	\centering
	\caption{Performance comparison among different methods for each human action via mean angle error.}
	\label{table_each1}
	\begin{center}
		{\scriptsize
			\begin{tabular}{|c|c|c|c|c|c|c|c|c|c|c|c|c|c|c|c|c|}
				\hline
				\multirow{3}{*}{Methods} &\multicolumn{8}{c|}{Directions} &\multicolumn{8}{c|}{Discussion}\\
				\cline{2-17}
				&	\multicolumn{4}{c|}{Short-term motion prediction}  &    \multicolumn{4}{c|}{Long-term motion prediction} & \multicolumn{4}{c|}{Short-term motion prediction} &\multicolumn{4}{c|}{Long-term motion prediction}\\
				\cline{2-17}
				&	\hspace{-0.55em}80ms\hspace{-0.55em}  &  \hspace{-0.55em} 160ms\hspace{-0.55em} &   \hspace{-0.55em} 320ms\hspace{-0.55em}& \hspace{-0.55em}400ms\hspace{-0.55em} &\hspace{-0.55em}    560ms\hspace{-0.5em} &\hspace{-0.55em}    640ms\hspace{-0.55em} &\hspace{-0.55em}    720ms\hspace{-0.5em} & \hspace{-0.55em}   1000ms \hspace{-0.55em}&\hspace{-0.55em}    80ms\hspace{-0.55em}&\hspace{-0.55em}     160ms \hspace{-0.55em}&\hspace{-0.55em}    320ms\hspace{-0.55em}&\hspace{-0.55em}     400ms\hspace{-0.55em}&\hspace{-0.55em}     560ms\hspace{-0.55em}&\hspace{-0.55em}     640ms \hspace{-0.55em}&\hspace{-0.55em}    720ms\hspace{-0.55em}&\hspace{-0.55em}     1000ms\\
				\hline
				ERD~\cite{fragkiadaki2015recurrent}   & 0.77  & 0.83  & 0.94  & 1.03  & 1.12  & 1.16  & 1.22  & 1.35  & 0.89  & 0.97  & 1.23  & 1.74  & 1.80   & 1.80   & 1.89  & 1.97 \\
				LSTM-3LR~\cite{fragkiadaki2015recurrent} & 0.60   & 0.75  & 0.82  & 0.97  & 1.06  & 1.10   & 1.17  & 1.23  & 0.81  & 0.90   & 1.12  & 1.64  & 1.75  & 1.67  & 1.85  & 1.93 \\
				Res-GRU~\cite{martinez2017human} & 0.32  & 0.52  & 0.77  & 0.89  & 0.96  & 1.04  & 1.09  & 1.13  & 0.36  & 0.70   & 1.01  & 1.09  & 1.56  & 1.65  & 1.75  & 1.91 \\
				MHU~\cite{tang2018long}   & 0.30   & {\bf 0.51}  & 0.76  & 0.84  & {0.91}   & 0.93  & {\bf 0.97} & 1.08  & {\bf 0.32}  & 0.68  & 0.95  & {\bf 1.02}  & 1.34  & 1.52  & 1.66  & 1.90 \\
				SC-RNN & {\bf 0.30}   & 0.52  & {\bf 0.69}  & {\bf 0.79}  & {\bf 0.87}  & {\bf 0.93}  & 1.01  & {\bf 1.07}  & 0.33  & {\bf 0.67}  & {\bf 0.94}  & 1.05  & {\bf 1.27}  & {\bf 1.49}  & {\bf 1.56}  & {\bf 1.86} \\
				
				\hline\hline
				\multirow{3}{*}{Methods} &\multicolumn{8}{c|}{Eating} &\multicolumn{8}{c|}{Greeting}\\
				\cline{2-17}
				&	\multicolumn{4}{c|}{Short-term motion prediction}  &    \multicolumn{4}{c|}{Long-term motion prediction} & \multicolumn{4}{c|}{Short-term motion prediction} &\multicolumn{4}{c|}{Long-term motion prediction}\\
				\cline{2-17}
				&	\hspace{-0.55em}80ms\hspace{-0.55em}  &  \hspace{-0.55em} 160ms\hspace{-0.55em} &   \hspace{-0.55em} 320ms\hspace{-0.55em}& \hspace{-0.55em}400ms\hspace{-0.55em} &\hspace{-0.55em}    560ms\hspace{-0.5em} &\hspace{-0.55em}    640ms\hspace{-0.55em} &\hspace{-0.55em}    720ms\hspace{-0.5em} & \hspace{-0.55em}   1000ms \hspace{-0.55em}&\hspace{-0.55em}    80ms\hspace{-0.55em}&\hspace{-0.55em}     160ms \hspace{-0.55em}&\hspace{-0.55em}    320ms\hspace{-0.55em}&\hspace{-0.55em}     400ms\hspace{-0.55em}&\hspace{-0.55em}     560ms\hspace{-0.55em}&\hspace{-0.55em}     640ms \hspace{-0.55em}&\hspace{-0.55em}    720ms\hspace{-0.55em}&\hspace{-0.55em}     1000ms\\
				\hline
				
				ERD~\cite{fragkiadaki2015recurrent}   & 0.73  & 0.78  & 0.84  & 0.93  & 1.02  & 1.07  & 1.11  & 1.15  & 1.07  & 1.33  & 1.67  & 1.84  & 2.39  & 2.31  & 2.34  & 2.39 \\
				LSTM-3LR~\cite{fragkiadaki2015recurrent} & 0.69  & 0.74  & 0.80   & 0.87  & 0.91  & 0.98  & 1.03  & 1.09  & 1.03  & 1.32  & 1.79  & 1.89  & 2.31  & 2.28  & 2.31  & 2.36 \\
				Res-GRU~\cite{martinez2017human} & 0.27  & 0.44  & 0.64  & 0.82  & 0.88  & 0.94  & 1.00    & 1.02  & 0.69  & 1.10   & 1.61  & 1.73  & 2.13  & 1.89  & 1.86  & 2.00 \\
				MHU~\cite{tang2018long}   & 0.25  & 0.46  & 0.61  & 0.72  & 0.77  & 0.81  & 0.87  & 0.90   & 0.56  & 0.91  & 1.36  & 1.51  & 1.79  & 1.77  & 1.82  & 1.88 \\
				SC-RNN & {\bf 0.25}  & {\bf 0.42}  & {\bf 0.56}  & {\bf 0.68}  & {\bf 0.67}  & {\bf 0.73}  & {\bf 0.80}   & {\bf 0.87}  & {\bf 0.54}  & {\bf 0.79}  & {\bf 1.13}  & {\bf 1.31}  & {\bf 1.49}  & {\bf 1.34}  & {\bf 1.63}  & {\bf 1.64} \\
				\hline\hline
				\multirow{3}{*}{Methods} &\multicolumn{8}{c|}{Phoning} &\multicolumn{8}{c|}{Posing}\\
				\cline{2-17}
				&	\multicolumn{4}{c|}{Short-term motion prediction}  &    \multicolumn{4}{c|}{Long-term motion prediction} & \multicolumn{4}{c|}{Short-term motion prediction} &\multicolumn{4}{c|}{Long-term motion prediction}\\
				\cline{2-17}
				&	\hspace{-0.55em}80ms\hspace{-0.55em}  &  \hspace{-0.55em} 160ms\hspace{-0.55em} &   \hspace{-0.55em} 320ms\hspace{-0.55em}& \hspace{-0.55em}400ms\hspace{-0.55em} &\hspace{-0.55em}    560ms\hspace{-0.5em} &\hspace{-0.55em}    640ms\hspace{-0.55em} &\hspace{-0.55em}    720ms\hspace{-0.5em} & \hspace{-0.55em}   1000ms \hspace{-0.55em}&\hspace{-0.55em}    80ms\hspace{-0.55em}&\hspace{-0.55em}     160ms \hspace{-0.55em}&\hspace{-0.55em}    320ms\hspace{-0.55em}&\hspace{-0.55em}     400ms\hspace{-0.55em}&\hspace{-0.55em}     560ms\hspace{-0.55em}&\hspace{-0.55em}     640ms \hspace{-0.55em}&\hspace{-0.55em}    720ms\hspace{-0.55em}&\hspace{-0.55em}     1000ms\\
				\hline
				ERD~\cite{fragkiadaki2015recurrent}   & 0.47  & 0.68  & 0.89  & 0.93  & 1.07  & 1.13  & 1.21  & 1.27  & 0.56  & 1.01 & 1.25  & 1.57  & 1.89  & 2.11  & 2.23  & 2.59 \\
				LSTM-3LR~\cite{fragkiadaki2015recurrent} & 0.39  & 0.52  & 0.74  & 0.90   & 1.01  & 1.07  & 1.15  & 1.23  & 0.57  & 1.02 & 1.45  & 1.91  & 1.93   & 2.16  & 2.38  & 2.67 \\
				Res-GRU~\cite{martinez2017human} & 0.27  & 0.46  & 0.69  & 0.82  & 0.95  & 1.04  & 1.10   & 1.16  & 0.41  & 0.79  & 1.11  & 1.28  & 1.81  & 1.77  & 2.26  & 2.57 \\
				MHU~\cite{tang2018long}   & 0.29  & 0.45  & 0.67  & 0.71  & 0.89  & 0.97  & 1.08  & 1.14  & {\bf 0.32}  & {\bf 0.62}  & 1.21  & 1.48  & {\bf 1.80}   & {\bf 2.08}  & {\bf 2.17}  & {\bf 2.51} \\
				SC-RNN & {\bf 0.27}  & {\bf 0.43}  & {\bf 0.62}  & {\bf 0.68}  & {\bf 0.79}  & {\bf 0.81}  & {\bf 1.01}  & {\bf 1.08}  & 0.34  & 0.67  & {\bf 1.20}   & {\bf 1.45}  & 1.86  & 2.11  & 2.20   & 2.53 \\
				
				\hline\hline				
				\multirow{3}{*}{Methods} &\multicolumn{8}{c|}{Purchases} &\multicolumn{8}{c|}{Sitting}\\
				\cline{2-17}
				&	\multicolumn{4}{c|}{Short-term motion prediction}  &    \multicolumn{4}{c|}{Long-term motion prediction} & \multicolumn{4}{c|}{Short-term motion prediction} &\multicolumn{4}{c|}{Long-term motion prediction}\\
				\cline{2-17}
				&	\hspace{-0.55em}80ms\hspace{-0.55em}  &  \hspace{-0.55em} 160ms\hspace{-0.55em} &   \hspace{-0.55em} 320ms\hspace{-0.55em}& \hspace{-0.55em}400ms\hspace{-0.55em} &\hspace{-0.55em}    560ms\hspace{-0.5em} &\hspace{-0.55em}    640ms\hspace{-0.55em} &\hspace{-0.55em}    720ms\hspace{-0.5em} & \hspace{-0.55em}   1000ms \hspace{-0.55em}&\hspace{-0.55em}    80ms\hspace{-0.55em}&\hspace{-0.55em}     160ms \hspace{-0.55em}&\hspace{-0.55em}    320ms\hspace{-0.55em}&\hspace{-0.55em}     400ms\hspace{-0.55em}&\hspace{-0.55em}     560ms\hspace{-0.55em}&\hspace{-0.55em}     640ms \hspace{-0.55em}&\hspace{-0.55em}    720ms\hspace{-0.55em}&\hspace{-0.55em}     1000ms\\
				\hline
				
				ERD~\cite{fragkiadaki2015recurrent}   & 0.76  & 0.89  & 0.98  & 1.12  & 1.26  & 1.35  & 1.40   & 1.57  & 0.44  & 0.93  & 1.24  & 1.59  & 1.74  & 1.96  & 2.01  & 2.10 \\
				
				LSTM-3LR~\cite{fragkiadaki2015recurrent} & 0.71  & 0.84  & 0.93  & 1.10   & 1.22  & 1.36  & 1.47  & 1.58  & 0.45  & 0.91  & 1.39  & 1.59  & 1.69  & 1.90   & 1.97  & 2.03 \\
				
				Res-GRU~\cite{martinez2017human} & 0.51  & 0.97  & 1.08  & 1.17  & 1.21  & 1.28  & 1.36  & 1.46  & 0.44  & 0.91  & 1.49  & 1.64  & 1.71  & 1.82  & 1.91  & 1.97 \\
				
				MHU~\cite{tang2018long}   & 0.48  & 0.96  & 1.09  & 1.12  & 1.18  & 1.24  & 1.30   & 1.37  & 0.44  & 0.89  & 1.12  & 1.34  & 1.56  & 1.67  & 1.78  & 1.84 \\
				
				SC-RNN & {\bf 0.47}  & {\bf 0.93}  & {\bf 1.07}  & {\bf 1.07}  & {\bf 1.13}  & {\bf 1.17}  & {\bf 1.24}  & {\bf 1.30}   & {\bf 0.43}  & {\bf 0.89}  & {\bf 1.12}  & {\bf 1.34}  & {\bf 1.50}   & {\bf 1.64}  & {\bf 1.76}  & {\bf 1.81} \\
				
				\hline\hline
				\multirow{3}{*}{Methods} &\multicolumn{8}{c|}{Sittingdown} &\multicolumn{8}{c|}{Smoking}\\
				\cline{2-17}
				&	\multicolumn{4}{c|}{Short-term motion prediction}  &    \multicolumn{4}{c|}{Long-term motion prediction} & \multicolumn{4}{c|}{Short-term motion prediction} &\multicolumn{4}{c|}{Long-term motion prediction}\\
				\cline{2-17}
				&	\hspace{-0.55em}80ms\hspace{-0.55em}  &  \hspace{-0.55em} 160ms\hspace{-0.55em} &   \hspace{-0.55em} 320ms\hspace{-0.55em}& \hspace{-0.55em}400ms\hspace{-0.55em} &\hspace{-0.55em}    560ms\hspace{-0.5em} &\hspace{-0.55em}    640ms\hspace{-0.55em} &\hspace{-0.55em}    720ms\hspace{-0.5em} & \hspace{-0.55em}   1000ms \hspace{-0.55em}&\hspace{-0.55em}    80ms\hspace{-0.55em}&\hspace{-0.55em}     160ms \hspace{-0.55em}&\hspace{-0.55em}    320ms\hspace{-0.55em}&\hspace{-0.55em}     400ms\hspace{-0.55em}&\hspace{-0.55em}     560ms\hspace{-0.55em}&\hspace{-0.55em}     640ms \hspace{-0.55em}&\hspace{-0.55em}    720ms\hspace{-0.55em}&\hspace{-0.55em}     1000ms\\
				\hline
				ERD~\cite{fragkiadaki2015recurrent}   & 0.35  & 0.99  & 1.67  & 1.80   & 1.89  & 1.97  & 2.13   & 2.18  & 0.49  & 0.47  & 1.05  & 1.18  & 1.32  & 1.49  & 1.51  & 1.56 \\
				LSTM-3LR~\cite{fragkiadaki2015recurrent} & 0.34  & 0.95  & {1.42}  & 1.75  & 1.82  & 1.96  & 2.07  & 2.14  & 0.42  & 0.45  & 0.94  & 1.12  & 1.20  & 1.31  & 1.35  & 1.48 \\
				Res-GRU~\cite{martinez2017human} & 0.34  & 0.97  & 1.52  & 1.68  & 1.79  & 1.92  & 2.01  & 2.05  & 0.36  & 0.39  & 1.06  & 1.13  & 1.19  & 1.22  & 1.29  & 1.32 \\
				MHU~\cite{tang2018long}   & 0.34  & {\bf 0.69}  & 1.38  & 1.53  & 1.66  & {\bf 1.72}  & {\bf 1.78}  & 1.93  & 0.35  & 0.42  & 0.96  & 1.08  & 1.12  & 1.17  & 1.24  & 1.30 \\
				SC-RNN & {\bf 0.34}  & 0.72  & {\bf 1.37}  & {\bf 1.51}  & {\bf 1.65}  & 1.75  & 1.81  & {\bf 1.92}  & {\bf 0.35}  & {\bf 0.41}  & {\bf 0.94}  & {\bf 1.06}  & {\bf 1.10}   & {\bf 1.15}  & {\bf 1.21}  & {\bf 1.28} \\
				
				\hline\hline
				\multirow{3}{*}{Methods} &\multicolumn{8}{c|}{Takingphoto} &\multicolumn{8}{c|}{Waiting}\\
				\cline{2-17}
				&	\multicolumn{4}{c|}{Short-term motion prediction}  &    \multicolumn{4}{c|}{Long-term motion prediction} & \multicolumn{4}{c|}{Short-term motion prediction} &\multicolumn{4}{c|}{Long-term motion prediction}\\
				\cline{2-17}
				&	\hspace{-0.55em}80ms\hspace{-0.55em}  &  \hspace{-0.55em} 160ms\hspace{-0.55em} &   \hspace{-0.55em} 320ms\hspace{-0.55em}& \hspace{-0.55em}400ms\hspace{-0.55em} &\hspace{-0.55em}    560ms\hspace{-0.5em} &\hspace{-0.55em}    640ms\hspace{-0.55em} &\hspace{-0.55em}    720ms\hspace{-0.5em} & \hspace{-0.55em}   1000ms \hspace{-0.55em}&\hspace{-0.55em}    80ms\hspace{-0.55em}&\hspace{-0.55em}     160ms \hspace{-0.55em}&\hspace{-0.55em}    320ms\hspace{-0.55em}&\hspace{-0.55em}     400ms\hspace{-0.55em}&\hspace{-0.55em}     560ms\hspace{-0.55em}&\hspace{-0.55em}     640ms \hspace{-0.55em}&\hspace{-0.55em}    720ms\hspace{-0.55em}&\hspace{-0.55em}     1000ms\\
				\hline
				
				ERD~\cite{fragkiadaki2015recurrent}   & 0.71  & 0.78  & 1.02  & 1.20   & 1.27  & 1.46  & 1.49  & 1.53  & 0.38  & 0.69  & 1.14  & 1.43  & 1.56  & 1.62  & 1.70   & 1.79 \\
				LSTM-3LR~\cite{fragkiadaki2015recurrent} & 0.74  & 0.64  & 0.91  & 1.19  & 1.29  & 1.40   & 1.45  & 1.51  & 0.35  & 0.65  & 1.13  & 1.41  & 1.52  & 1.60   & 1.68  & 1.77 \\
				Res-GRU~\cite{martinez2017human} & 0.31  & 0.56  & 0.87  & 1.04  & 1.18  & 1.25  & 1.19  & 1.37  & 0.36  & 0.62  & 1.15  & 1.40   & 1.47  & 1.58  & 1.69  & 1.72 \\
				MHU~\cite{tang2018long}   & 0.29  & 0.57  & 0.84  & 0.98  & 1.02  & 1.08  & 1.16  & 1.34  & 0.32  & 0.56  & 1.12  & 1.28  & 1.36  & 1.49  & 1.57  & 1.68 \\
				SC-RNN & {\bf 0.25}  & {\bf 0.53}  & {\bf 0.81}  & {\bf 0.90}   & {\bf 0.97}  & {\bf 1.02}  & {\bf 1.12}  & {\bf 1.27}  & {\bf 0.30}   & {\bf 0.54}  & {\bf 1.11}   & {\bf 1.23}  & {\bf 1.31}  & {\bf 1.40}   & {\bf 1.52}  & {\bf 1.63} \\
				
				\hline\hline
				\multirow{3}{*}{Methods} &\multicolumn{8}{c|}{Walking} &\multicolumn{8}{c|}{Walkingtogether}\\
				\cline{2-17}
				&	\multicolumn{4}{c|}{Short-term motion prediction}  &    \multicolumn{4}{c|}{Long-term motion prediction} & \multicolumn{4}{c|}{Short-term motion prediction} &\multicolumn{4}{c|}{Long-term motion prediction}\\
				\cline{2-17}
				&	\hspace{-0.55em}80ms\hspace{-0.55em}  &  \hspace{-0.55em} 160ms\hspace{-0.55em} &   \hspace{-0.55em} 320ms\hspace{-0.55em}& \hspace{-0.55em}400ms\hspace{-0.55em} &\hspace{-0.55em}    560ms\hspace{-0.5em} &\hspace{-0.55em}    640ms\hspace{-0.55em} &\hspace{-0.55em}    720ms\hspace{-0.5em} & \hspace{-0.55em}   1000ms \hspace{-0.55em}&\hspace{-0.55em}    80ms\hspace{-0.55em}&\hspace{-0.55em}     160ms \hspace{-0.55em}&\hspace{-0.55em}    320ms\hspace{-0.55em}&\hspace{-0.55em}     400ms\hspace{-0.55em}&\hspace{-0.55em}     560ms\hspace{-0.55em}&\hspace{-0.55em}     640ms \hspace{-0.55em}&\hspace{-0.55em}    720ms\hspace{-0.55em}&\hspace{-0.55em}     1000ms\\
				\hline
				ERD~\cite{fragkiadaki2015recurrent}   & 0.32  & 0.56  & 1.01  & 1.15  & 1.21  & 1.28  & 1.36  & 1.43  & 0.43  & 0.67  & 0.78  & 0.89  & 1.27  & 1.34  & 1.42  & 1.56 \\
				LSTM-3LR~\cite{fragkiadaki2015recurrent} & 0.35  & 0.56  & 0.89  & 1.07  & 1.19  & 1.27  & 1.33  & 1.37  & 0.42  & 0.67  & 0.89  & 0.91  & 1.19  & 1.29  & 1.34  & 1.42 \\
				Res-GRU~\cite{martinez2017human} & 0.30   & 0.48  & 0.70   & 0.76  & 0.93  & 0.96  & 1.02  & 1.07  & 0.40   & 0.63  & 0.88  & 0.97  & 1.12  & 1.24  & 1.29  & 1.35 \\
				MHU~\cite{tang2018long}   & 0.34  & 0.54  & 0.69  & 0.78  & 0.89  & 0.98  & 1.07  & 1.12  & 0.41  & 0.55  & 0.61  & 0.79  & 0.97  & 1.08  & 1.16  & 1.27 \\
				SC-RNN & {\bf 0.32}  & {\bf 0.47}  & {\bf 0.54}  & {\bf 0.63}  & {\bf 0.72}  & {\bf 0.78}  & {\bf 0.83}  & {\bf 0.88}  & {\bf 0.41}  & {\bf 0.43}  & {\bf 0.50}   & {\bf 0.58}  & {\bf 0.67}  & {\bf 0.73}  & {\bf 0.89}  & {\bf 1.02} \\
				\hline
			\end{tabular}
		}
	\end{center}
\end{table*}

\begin{figure*}[!t]
	\vspace{3mm}
	\centering
	\includegraphics[scale=0.18]{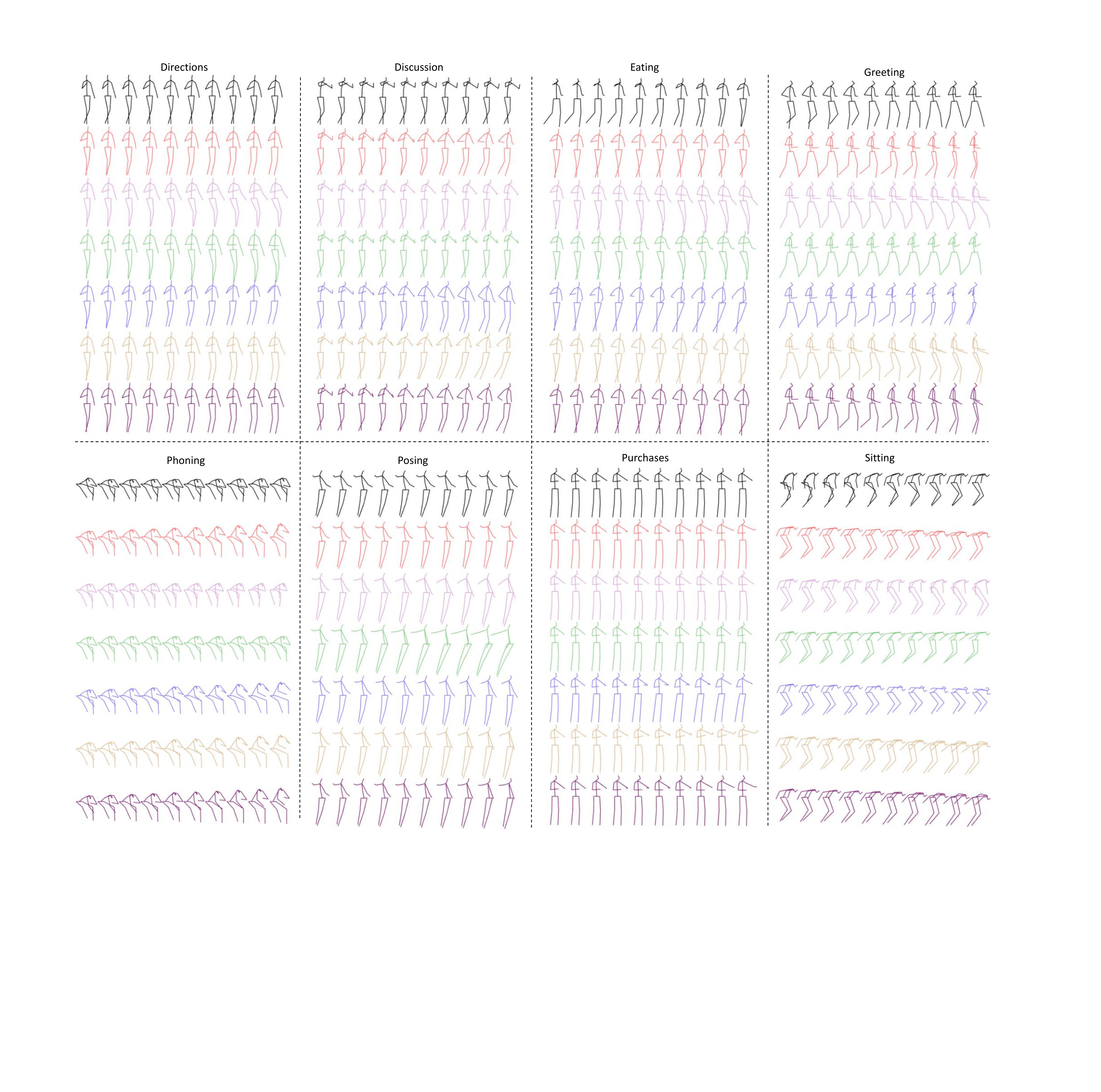}
	\caption{Some visualized results of the motion prediction for all 15 actions. In each group, the first and the second rows are the observed motion and the ground-truth human motions, respectively; and the 3rd-7th rows are the human motion predicted by ERD~\cite{fragkiadaki2015recurrent}, LSTM-3LR~\cite{fragkiadaki2015recurrent}, Res-GRU~\cite{martinez2017human}, MHU~\cite{tang2018long} and the proposed SC-RNN, respectively. For better view, please see $\times3$ original color PDF. This figure is followed by Figure~\ref{show_2}.}
		\vspace{2.5mm}
	\label{show_1}
\end{figure*}

\begin{figure*}[!t]
	\vspace{3mm}
	\centering
	\includegraphics[scale=0.18]{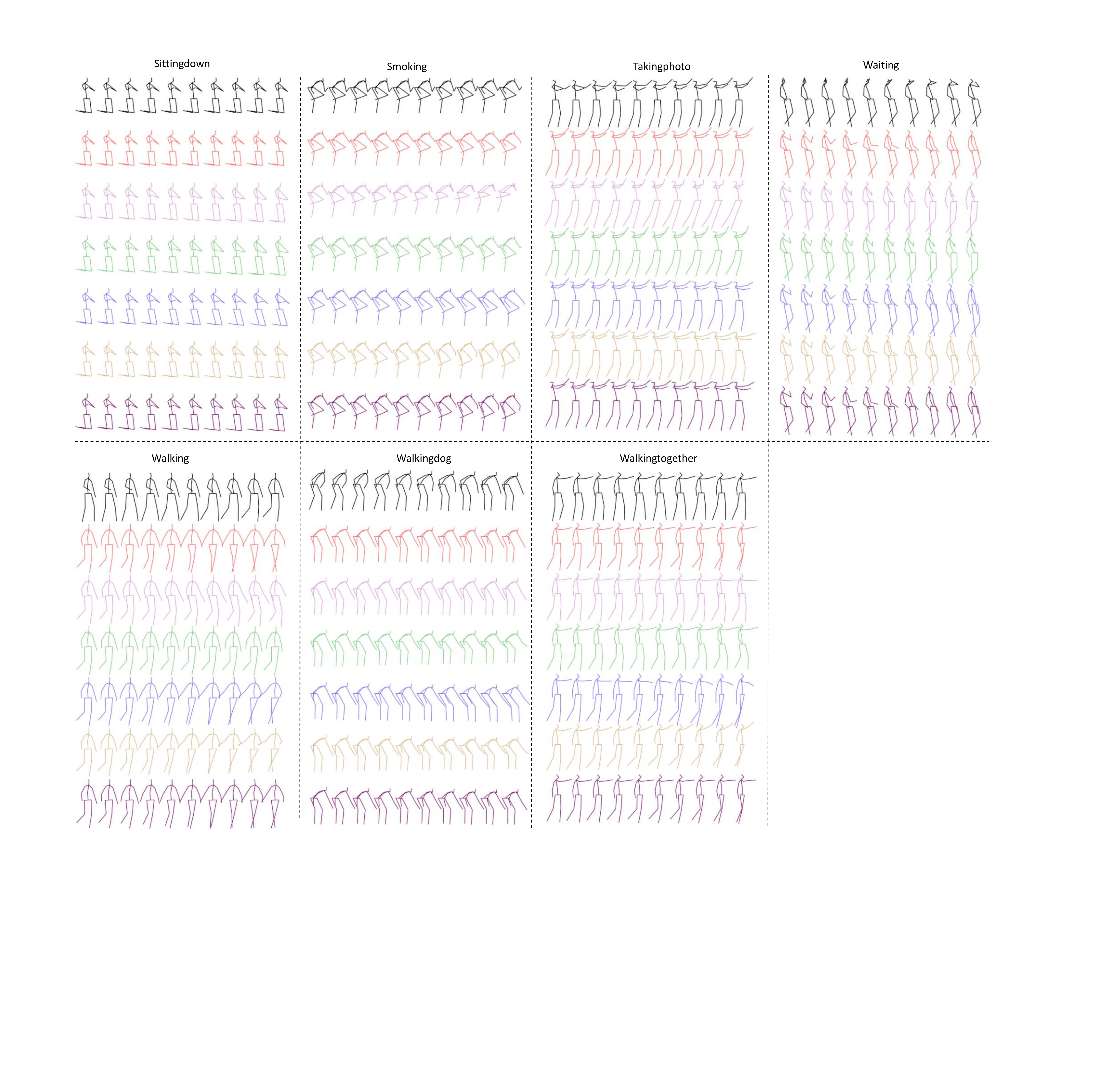}
	\caption{Some visualized results of the motion prediction for all 15 actions. In each group, the first and the second rows are the observed motion and the ground-truth human motions, respectively; and the 3rd-7th rows are the human motion predicted by ERD~\cite{fragkiadaki2015recurrent}, LSTM-3LR~\cite{fragkiadaki2015recurrent}, Res-GRU~\cite{martinez2017human}, MHU~\cite{tang2018long} and the proposed SC-RNN, respectively. For better view, please see $\times3$ original color PDF. This figure follows Figure~\ref{show_1}.}
	\label{show_2}
	\vspace{2.5mm}
\end{figure*}

\subsection{Comparison with the competitive methods} To illustrate the superiority of the proposed SC-RNN, we compare SC-RNN with several related methods, including Encoder-Recurrent-Decoder (ERD)~\cite{fragkiadaki2015recurrent},  3-Layer Long Short-Term Memory cells) (LSTM-3LR)~\cite{fragkiadaki2015recurrent}, Residual Gated Recurrent Unit (Res-GRU)~\cite{martinez2017human}, and Modified High-way Unit (MHU)~\cite{tang2018long}. Table~\ref{table_results} shows the MAEs achieved by different methods on the H3.6M dataset. If can be seen that SC-RNN performs better than the alternatives, especially ERD and LSTM-3LR. In particular, SC-RNN has gained about 0.8 reduction compared with the MHU in terms of the long-term (1000ms) human motion prediction. This illustrates that SC-RNN is more effective than the other methods for predicting the future motion based on the observed human motion sequence.

We also compare the MAEs obtained by different methods on each human action in the H3.6M dataset, as shown in Table~2. SC-RNN achieves lowest MAE on most of the human actions.  We can see that: 1) for the ``eating" action, all the methods achieves satisfactory performance; 2) for the ``posing" action, the MAEs on the long-term (1000ms) achieved by all methods are larger than $2.5$, since the motions of different persons are largely diverse from each other; 3) for the ``greeting" action, the proposed SC-RNN makes significant improvement than the other methods, since SC-RNN explores the spatial coherence among joints, which is crucial for modeling the human motion with large variation of the arm and leg motions. Although SC-RNN does not achieve the lowest MAE on the ``posing" action, it is comparable to the state-of-the-art method (MHU). 

We also qualitatively compare the proposed SC-RNN with the state-of-the-art methods, as shown in Figure~\ref{show_1}. SC-RNN generates the authentic motions for these actions, most of which are pretty closely to the ground-truth motions. For the human motions generated by ERD, LSTM-3LR, and Res-GRU, some details are not accurate compared with the ground truth, such as leg joints in the ``eating" action. Since MHU considers all observed motions as motion context to predict the human motion, it can generate more reliable motions than ERD and LSTM-3LR. SC-RNN can predict motions with high accuracy in some details. For example, in the ``greeting" action, the locations of two hands predicted by SC-RNN are more accurate than MHU in the long-term (1000ms) motion prediction.

\vspace{3mm}
\section{Conclusions}
\label{C}
In this work, for human motion prediction, we propose a novel Skeleton-joint Co-attention Recurrent Neural Networks (SC-RNN) to simultaneously capture the spatial coherence among joints and the temporal evolution among skeletons on a co-attention feature map, by utilizing all useful historical motion information
as the motion context. The architecture of the proposed SC-RNN has three advantages. First, a new Skeleton-joint Co-Attention (SCA) is designed to dynamically learn a co-attention feature map on the skeleton-joint feature map of all the observed motions over time, which can refine the useful observed motion information by the different skeleton-joint attention factors. Second, a Skeleton-joint Co-attention GRU (SC-GRU) embedded with SCA is proposed to model human-joint motions and human-skeleton motions simultaneously in spatiotemporal space. Third, a new weighted gram-matrix loss is presented to train the SC-GRU model in spatiotemporal space. Experimental results on human motion prediction well demonstrate that the proposed SC-RNN outperforms the related methods.


\vspace{3mm}

\bibliographystyle{IEEEtran}
\bibliography{IEEEabrv,egbib}

\end{document}